# OneTrack-M - A multitask approach to transformer-based MOT models

Luiz Carlos Silva de Araujo  [ Federal University of Amazonas | *luizcls@icomp.ufam.edu.br* ]
Carlos Mauricio Seródio Figueiredo  [ State University of Amazonas | *cfigueiredo@uea.edu.br* ]

**Abstract** Multi-Object Tracking (MOT) is a critical problem in computer vision, essential for understanding how objects move and interact in videos. This field faces significant challenges such as occlusions and complex environmental dynamics, impacting model accuracy and efficiency. While traditional approaches have relied on Convolutional Neural Networks (CNNs), introducing transformers has brought substantial advancements. This work introduces OneTrack-M, a transformer-based MOT model designed to enhance tracking computational efficiency and accuracy. Our approach simplifies the typical transformer-based architecture by eliminating the need for a decoder model for object detection and tracking. Instead, the encoder alone serves as the backbone for temporal data interpretation, significantly reducing processing time and increasing inference speed. Additionally, we employ innovative data pre-processing and multitask training techniques to address occlusion and diverse objective challenges within a single set of weights. Experimental results demonstrate that OneTrack-M achieves at least 25% faster inference times compared to state-of-the-art models in the literature while maintaining or improving tracking accuracy metrics. These improvements highlight the potential of the proposed solution for real-time applications such as autonomous vehicles, surveillance systems, and robotics, where rapid responses are crucial for system effectiveness.

## 1 Introduction

Computer vision simulates the complex nature of human visual perception, enabling machines to interpret visual data. This field continuously strives to replicate the human ability to process and understand images, leading to the development of advanced technologies. Significant progress has been driven by algorithmic enhancements and powerful hardware, such as advanced CPUs and GPUs, which facilitate the processing of large data volumes and the execution of complex machine learning (ML) and deep learning (DL) models.

Among computer vision applications, Multi-Object Tracking (MOT) Bashar *et al*. [2022] stands out as crucial for understanding how objects move and interact in videos. This domain faces challenges like occlusions and complex environmental dynamics, traditionally addressed by a progression of models from feature-based methods to advanced ML algorithms.

In computer vision, Convolutional Neural Networks (CNNs) have been foundational for object detection and classification, tasks essential for MOT Wojke *et al*. [2017]. CNNs efficiently capture visual features, from simple edges to complex patterns. However, introducing transformers has marked significant advancements, outperforming CNNs in tasks like object classification, such as the results in Dosovitskiy *et al*. [2021], and video processingLiu *et al*. [2021b]. This is due to the transformers' ability to model long-range relationships in data, making them ideal for understanding contexts and temporal relationships between objects.

Originally developed for natural language processing, transformers also have shown substantial potential in MOT applications, offering new possibilities for analyzing and interpreting complex image sequences. Despite their promise, transformer models still face challenges related to efficiency, both computationally and in tracking accuracy, prompting the search for innovative solutions.

Computational efficiency remains a common challenge in MOT, encouraging the exploration of advanced ML/DL techniques combined with hardware optimizations to enhance tracking accuracy while reducing computational resource requirements. Some of the main works that use transformers in MOT have major speed issues, such as Zeng *et al*. [2022], Sun *et al*. [2021], and Meinhardt *et al*. [2022]. But not only them, as research in this field tends to favor accuracy over inference speed, meaning most models will find issues when applied to real-world scenarios where skipping frames can become an issue.

The necessity for research on reducing inference times in MOT systems is underscored by applications that depend on real-time processing, such as autonomous vehicles, which rely on rapid responses to avoid obstacles and prevent collisions, and surveillance systems, which require quick processing to detect anomalies and track individuals effectively. Swift model responses are crucial in robotics and other fields requiring real-time environmental interpretation by intelligent agents. This work introduces an approach to enhancing efficiency with transformers in MOT systems by leveraging transformers to improve system accuracy while maintaining practical inference speeds. Our approach simplifies a common step in transformer-based models by eliminating the need for a decoder model for generating object detections and tracks, with the encoder alone serving as the backbone for temporal data interpretation. This significantly reduces processing time and increases speed compared to similar models. Additionally, we address a common issue in end-to-end models—very different objectives with a single set of weights being trained—by incorporating multitasking training techniques and concepts.

The contributions of this work are as follows:

- OneTrack-M represents a method to build and train such MOT models, now emphasizing achieving faster inference times.
- The training technique used is innovative, as most works will try to train the model by adjusting the weights of losses to form a composite loss.
- The channel-wise encoding method used was developed to reflect how our input data was modeled.

This article is structured as follows: Section 2 presents a critical analysis of the field, including both transformer and non-transformer methods, forming the basis for our decisions. Section 3 describes the methods and models developed for this project. Section 4 details experiments and results, comparing them with other MOT models. Finally, Section 5 concludes the discussion and suggests directions for future work.

## 2 Related Works

This chapter covers the main works that influenced the decisions made in OneTrack-M design.

### 2.1 Classical ML Approach to MOT

First, it's important to clarify that "classical" refers to machine learning techniques that are not considered deep, such as Support Vector Machines (SVMs), K-nearest neighbors, K-means, and others. Although these techniques lack the generalization capabilities of deep methods, they still have value for solving simpler problems with less data and requiring shorter inference times.

In the case of MOT, which is a complex problem, the classical techniques that show good results are mainly those based on the Kalman filter, such as [Bewley *et al.*, 2016]. In this case, the Kalman filter provides temporal context for the system, working sequentially with any object detection model. The idea is to predict the future position from previous detections in a window of frames. From these values, the current detections are associated with the previous ones. More generally, particle filters are also an option to infer the trajectory of objects over time, as done in Jaward *et al.* [2006].

Another approach to this is optical flow to model the movement of each object over time, as in Su *et al.* [2019]. Optical flow techniques calculate the motion of objects by analyzing the differences in intensity between consecutive frames. This method helps in tracking by providing motion vectors, which can be used to predict the next positions of the objects, aiding in associating detections across frames. Optical flow is particularly useful in scenarios where objects move continuously and smoothly, making it a valuable tool in the MOT toolkit.

### 2.2 Deep Learning in MOT

Recent advances in deep learning techniques have revolutionized the field of Multi-Object Tracking (MOT), overcoming the limitations of traditional and classical approaches, particularly in challenging scenarios. Models like DeepSORT Wojke *et al.* [2017], which combines the Kalman filter with features extracted by deep networks, exemplify the state-of-the-art in MOT, showcasing the superior generalization capabilities of these models.

Introducing deep learning into MOT broke down previous barriers, enabling various applications and innovations. In this context, Convolutional Neural Networks (CNNs) stand out for their ability to extract local information from images, widely adopted in deep learning models as shown in works such as Du *et al.* [2023], Zhang *et al.* [2022], and Meinhardt *et al.* [2022]. In these works, the convolutional part of the architecture is responsible for extracting image features that are processed in a subsequent stage.

In addition to spatial features, processing temporal sequences is essential for tracking objects in videos. Recurrent Neural Networks (RNNs) are used to maintain relevant information over time, while transformers, with their attention layers, are emerging as a robust approach to relate image sequences, as illustrated in Zeng *et al.* [2022] and Meinhardt *et al.* [2022].

Beyond temporal features, re-identification-based methods use objects' visual aspects to associate identities, as seen in Aharon *et al.* [2022] and Mostafa *et al.* [2022]. Although efficient, these methods can lose crucial temporal information, as they rely solely on the similarity of objects over different frames, rather than modeling for their movement or even both simultaneously.

CenterTrack Zhou *et al.* [2020] introduced a new way of solving MOT. It innovates by rethinking how data is modeled, making the association process between frames a matter of tracking each centroid displacement across frames. This approach makes inference more direct and changes how objects are detected by using heatmaps. Thus, the model can learn all these parts in a unified way.

Finally, it is essential to highlight that the progress of MOT is closely linked to the advances in object detection brought by deep learning. This synergy between detection and tracking improves existing models and paves the way for developing new techniques and approaches in MOT.

### 2.3 Image Models with Transformers

Transformers improved natural language processing, and now they play a crucial role in computer vision by providing innovative approaches to complex tasks. The Vision Transformer (ViT) Dosovitskiy *et al.* [2021] adopts a novel strategy by splitting images into fixed-size patches, treating them as sequences of tokens, and applying attention layers to capture global dependencies. This approach contrasts with the local convolutions of CNNs, enabling ViT, when trained with large data and parameters, to outperform CNN models in image classification.

However, new transformer variants, such as the Swin Transformer Liu *et al.* [2021a], advance even further in computer vision. The Swin Transformer excels in its ability to efficiently model image hierarchies, adapting to different scales and contexts, making it particularly suitable for object detection and segmentation tasks. This model even serves as the basis for another state-of-the-art model in video classification: the Video Swin Transformer Liu *et al*. [2021b].

In parallel, DETR (Detection Transformer) Carion *et al*. [2020] represents another significant innovation, simplifying the object detection process by eliminating complex post-processing steps like Non-Maximum Suppression (NMS). DETR models object detection as a set prediction problem, using a combination of transformer encoder and decoder to generate direct detections without predefined anchors.

These innovations in transformer-based models demonstrate a significant technical advance and expand the possibilities for developing more efficient and accurate MOT systems. Using transformers in MOT promises a new wave of approaches that can better handle challenges such as occlusions and variations in object size, enhancing tracking system robustness and effectiveness.

Thus, transformers bring new techniques and methodologies to MOT while establishing a new paradigm in processing and interpreting visual data, paving the way for continuous innovations in multi-object tracking.

## 2.4 Transformers in Multi-Object Tracking (MOT)

Incorporating transformer-based models into Multi-Object Tracking (MOT) represents a paradigmatic shift, offering innovative and sophisticated approaches to traditional challenges in this field. One pioneering example is the TrackFormer model described in Meinhardt *et al*. [2022]. This model employs an encoder-decoder transformer architecture, similar to those used in Natural Language Processing (NLP) tasks, to track objects in video sequences. The distinctive aspect of TrackFormer lies in its tracking methodology: it continuously generates and updates tracking queries to maintain object identification across frames. This approach demonstrates transformers' ability to manage complex spatial and temporal information, maintaining tracking consistency without needing a dedicated object detection model.

The MOTR model introduced by Zeng *et al*. [2022] represents another significant advancement. MOTR expands the concept of DETR to the MOT context, adopting an encoder-decoder structure to detect and track objects simultaneously. It uses tracking queries that update over time, similar to TrackFormer but with a distinct implementation. This technique allows MOTR to handle object detection as a set prediction problem, improving tracking accuracy and computational efficiency. This innovation marks an important milestone in simplifying the MOT process and enhancing its effectiveness.

Additionally, the development of TransTrack, as reported by Sun *et al*. [2021] also significantly contributed to MOT. TransTrack adopts a hybrid approach, integrating convolutions for feature extraction and a transformer for detection association. This hybrid model illustrates the potential of combining CNNs and transformers in MOT, effectively addressing tracking challenges in complex contexts and suggesting a promising direction for future research.

The aforementioned models indicate substantial progress in MOT methods, demonstrating how transformers can reshape tracking strategies, leading to more accurate, efficient, and robust systems. Successfully integrating transformers into MOT signals a considerable advance in computer vision, pointing to new possibilities for research and development. However, two common issues exists in all these models: the use of a complete transformer architecture (i.e., encoder and decoder) and very high inference times, making real-time usage impractical.

## 2.5 Limitations of Current Solutions

The main limitation found in most recent works, which have been a focus of development in the field, is the inference time. Works such as Mostafa *et al*. [2022] and Zhang *et al*. [2021] have simplified the system into a single model that performs both detection and tracking end-to-end. Both of these employ a CNN to extract features from the images and then perform MOT. However, they do not excel in metrics evaluation when compared to other slower methods.

As previously mentioned, these faster approaches use CNNs as the backbone of their solution. These methods lack in both fields—metrics and speed—issues that are addressed by the solution proposed in OneTrack-M. This improvement is largely made possible by an observation made by the authors of Zhang *et al*. [2021], who describe an unfairness regarding the training step of such end-to-end models.

## 2.6 Multitask Training (MTL)

Regarding the effectiveness of multitask training, in Caruana [1997], the authors show that neural networks trained on multiple related tasks can achieve better performance in each task compared to isolated training, proving that the method can lead to better knowledge transfer and optimal weight selection to solve a multifaceted problem.

Some studies suggest that MTL can be effective in data-scarce scenarios since joint learning enables the model to use information from all tasks to improve its generalization capability. In Zhang and Yang [2017], the authors explore approaches and discuss how the method can overcome data scarcity.

This approach closely resembles how authors develop and train their MOT models, where an end-to-end model is commonly trained with all tasks simultaneously. However, Zhang *et al*. [2021] argues about the dissonance between learning these tasks, which harms the model's overall learning, such that training one task hinders the learning of another. The authors' solution was to use the joint optimization approach to solve this issue, with specific loss functions for each task. In our case, we propose a novel approach, using multi-step training, where the model is trained more than once, each time with one target task (first detection and then tracking), but all in the same model. More details about this approach are given in Section 3.6.1.

## 3 OneTrack-M Model

The development of OneTrack-M relies significantly on the studies by Zhou *et al*. [2020] and Zeng *et al*. [2022]. The former establishes a reference model that uses object centers for tracking, while the latter serves as a benchmark for tracking approaches that integrate transformers. Both works provide a theoretical and methodological foundation for this project.

The OneTrack-M architecture incorporates the pre-trained encoder from the model Dosovitskiy *et al.* [2021], which is used to extract features from image sequences and generate attention maps. This methodological choice leverages the Vision Transformer's effectiveness in understanding visual nuances present in image sequences, a crucial capability for efficient multi-object tracking.

Specific layers are added on top of this model for key multi-object tracking (MOT) functions, detailed below:

- **Heatmap Head**: Identifies object centers in each image.
- **Dimension Head**: Estimates the dimensions (height and width) of detected objects.
- **Center Displacement Head:** Calculates the variations in object center positions between frames.

Figure 1 illustrates the complete architecture, showing the interconnection of these components. The following sections better describe each element, clarifying their functions and the role they play in the OneTrack-M model's overall context.

## 3.1 Preprocessing

The initial stage of our model involves specific data handling to present it appropriately to the neural network. The main challenge here is to provide temporal context between selected frames. This is done by stacking all images from a temporal window along the channel dimension. For a standard image with dimensions $[3, Width, Height]$, considering a window size $W$, the input format becomes $[3 * W, Width, Height]$. From these images, patches are generated and later converted into tokens via a linear layer.

This methodology allows the neural network to fully absorb the temporal context, aligning with the preprocessing strategies adopted in previous works like Dosovitskiy *et al.* [2021], but adapting it for video analysis in this project. The components of this model section are illustrated in Figure 2. It is worth noting that no input is needed for the network besides the images, normalized and resized into a square matrix of size 224, which is the same as that used in the reference ViT model.

### 3.1.1 Image Stack and Patch Generation

According to the procedure described by the Vision Transformer authors, this stage consists of, given a square size (16x16 or 32x32), equally dividing the image into patches. This work maintained the 224x224 image size, again compatible with the experimental procedure of the previously mentioned base work. Thus, for 16x16 size patches, 196 small images are obtained $(\frac{224}{16})^2$. Each one passes through a linear layer to generate a vector projection of each patch, with a size of 768.

This stage becomes efficient because it uses a convolutional layer with an output of 768 channels, a kernel, and a stride matching the patch size. Thus, each convolution step processes a patch, either 16x16 or 32x32, without overlap, and then projects the layers to the desired size, meaning these square patches become a 768-sized vector. In Figure 1 these steps are illustrated in the Stacked Images, Stack Patch, and Linear Projection blocks.

### 3.1.2 Normalizations

In the context of tracking, which relies on analyzing object center displacements, and considering a video stream at 60 frames per second, a 5-frame temporal window represents only 83 milliseconds. The displacements observed between consecutive frames tend to show relatively small magnitudes in this short interval. To enhance the model's predictive ability given this characteristic, displacement values were normalized to the range $[-1, 1]$. The normalization parameters were determined through an exploratory analysis of the benchmark on the dataset MOT17. If the model is adapted to different contexts or datasets, it's necessary to revisit and adjust the normalization values considering the new data's displacement peculiarities.

The specific values used in the normalization are detailed in Table 1, providing a quantitative reference for this crucial preprocessing stage. This practice not only makes data interpretation easier for the model but also ensures more efficient adaptation to subtle movement variations, enhancing tracking accuracy.

**Table 1.** Displacements observed in the MOT17 training dataset Milan *et al.* [2016].

| Min. X | Max. X | Min. Y | Max. Y |
|---|---|---|---|
| -0.0174 | 0.0057 | -0.0157 | 0.0166 |

## 3.2 Channel Wise Encoding

Despite the effort to incorporate temporal context, the model may face difficulties due to the lack of a clear indication of the sequential order of image patches. Therefore, providing explicit positional context is crucial to optimize performance in attention-based models.

In this scenario, we opted for an innovative approach through the implementation of a technique called Channel Wise Encoding. This method doesn't limit itself to simple, fixed, or learned positional encodings but seeks a more comprehensive representation of the data's spatial and temporal structure. Specifically, it adds unique embeddings for each image patch channel, reflecting the temporal window in video processing.

Channel Wise Encoding innovates by assigning a distinct embedding vector for each frame within the temporal window, distinguishing not only the spatial features but also the temporal position of each patch within the sequence. This approach is vital for the model to understand object movement and order, as the sequential image sequence provides crucial tracking information.

Initially, these encodings are a trainable parameter tensor of dimensions $[W, n_d]$, where $W$ is the temporal window size and $n_d$ is the embedding dimension. This vector is then added to the vector derived from the image patches. Consequently, this strategy not only facilitates differentiation between frames but also improves the model's ability to capture temporal dynamics. It represents an effective solution to integrating temporal information into transformer-based models designed for multi-object tracking, with the additional benefit of minimizing inference time impact.

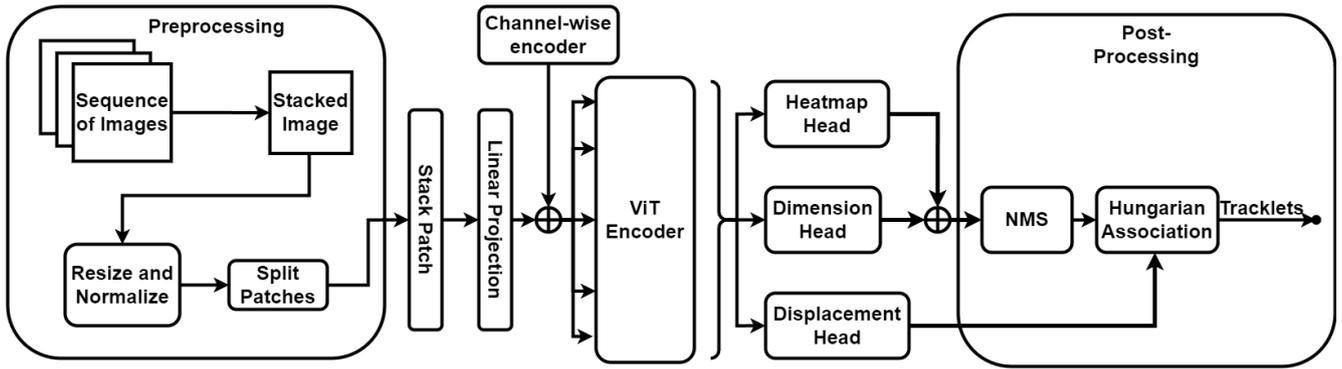

**Figure 1.** Implemented architecture for the OneTrack-M model.

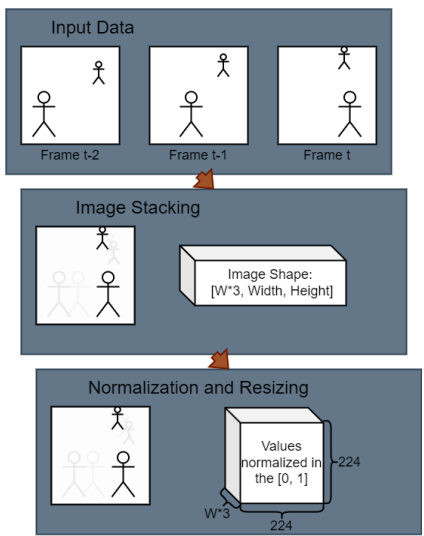

**Figure 2.** Detailed representation of preprocessing step for the model.

These vectors are added to the linear projections of each patch obtained in the previous preprocessing stage.

### 3.3 The ViT Encoder

The strategy adopted by the model involves using the Vision Transformer encoder pre-trained on ImageNet Murad and Pyun [2017] to extract attention maps, converting each image patch into an input sequence element. Unlike the conventional application of ViT for classification, the classification token is omitted in this context. This modification adapts ViT to function as a feature extractor focused on understanding spatial and temporal context while retaining the original settings of the pre-trained network.

This approach is significantly different from traditional methods described in literature, which often employ an encoder-decoder architecture, as evidenced in works like Sun *et al.* [2021], Meinhardt *et al.* [2022], Zeng *et al.* [2022], and Zhang *et al.* [2023]. Although robust, these solutions face the challenge of long inference periods. By simplifying to just the encoder and transferring the responsibility of the final output to a more efficient component, OneTrack-M has managed to significantly reduce processing time.

The choice of a pre-trained network is justified by the large data volume required for effective training of transformer networks. The limitation of the available data in the benchmark dataset used, combined with concerns that including external data could compromise comparability between different models, reinforces the relevance of this decision. Thus, using the pre-trained ViT emerges as a pragmatic solution, allowing prior learnings to be leveraged without requiring extensive training sets, aligning with OneTrack-M's goals of efficiency and effectiveness.

### 3.4 Output Heads

OneTrack-M's output structure consists of three main components: heatmaps, dimensions, and displacements. As shown in Figure 3, each of these heads comprises a sequence of two fully connected layers followed by two convolutional layers, specifically activated for each output type. For heatmaps and object dimensions, sigmoid activation is used, while hyperbolic tangent activation is necessary for displacements to represent the $[-1, 1]$ value range possible for this output.

The heatmap processing stage that adjusts its format to that expected by the output heads is illustrated in Figure 4. For this stage, we will process the attention maps into a lower dimension using a series of fully connected layers, resulting in a tensor of shape [batch, 196*W, 196], which gets reshaped into [batch, W, 196, 196]. This final tensor is the input used to perform inference in the three implemented heads.

This configuration aims to detect information grids, aligning with the output method proposed by CenterTrack Zhou *et al.* [2020]. Differentiation lies in the occlusion handling: unlike the baseline, which is limited to two consecutive images, our approach stacks multiple frames for input. This allows the model to preserve the displacement context of objects for a longer period, even in the presence of occlusions.

#### 3.4.1 Heatmap Head

The heatmap output identifies object centers, allowing unlimited object tracking—a significant advantage over other MOT approaches with transformers, like Zeng *et al.* [2022] and Zhang *et al.* [2023]. To manage complexity, the image dimension is reduced before inference and then expanded back to its original dimensions. This procedure is consistent across all model heads. The activation for this output is done by a sigmoid function, limiting values between 0 and 1.

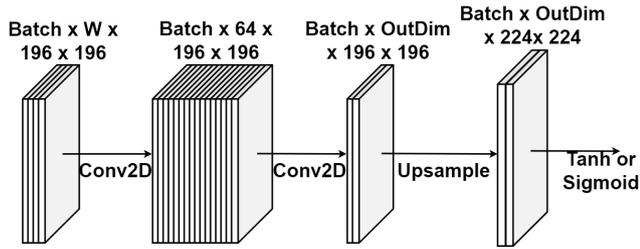

**Figure 3.** Detailed schematic of the network responsible for the last part of the model's heads. Each head has the same structure, varying only in activation function and output dimension. For heatmaps, $OutDim = 1$, while for the other two, $OutDim = 2$.

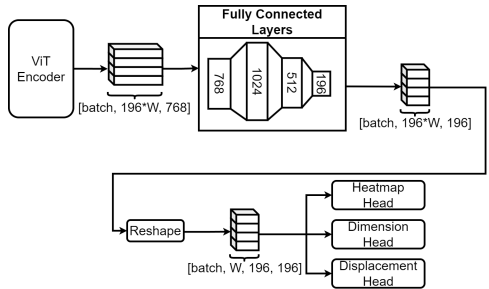

**Figure 4.** Output stage schematic after data is processed by Vision Transformer. The first tensor block represents attention maps extracted by the model. Each block is a sub-step of this final model stage.

Figure 5 compares a dataset sample with a model prediction, where each object is marked by a peak in the heatmap, smoothed by a Gaussian filter as shown by Zhou *et al.* [2020] and Zhou *et al.* [2019].

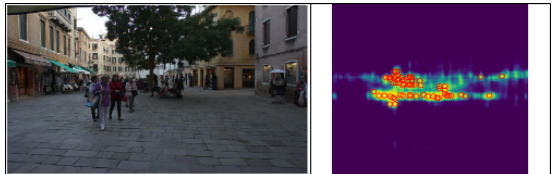

**Figure 5.** Example of an image with heatmap detection and its peaks. A concentration of people is noticeable in the central region, with empty areas in the upper and lower regions, demonstrating good model capability.

#### 3.4.2 Dimension Head

This head estimates the dimensions (width and height) of each object using a grid of shape $[2, Width, Height]$. Each matrix element corresponds to an object's dimensions at its central location. Although the outputs are processed simultaneously, fine-tuning is necessary to integrate these results into the final inference. This output is activated by a sigmoid function to keep the range between 0 and 1. This head is similar to the one implemented in CenterTrack Zhou *et al.* [2020].

#### 3.4.3 Displacement Head

Working similarly to the Dimension Head but with different weights, this part of the model calculates object center displacements in the two image dimensions, considering the previous frame. Even with data from earlier instances, the regression is only made between the current and the immediate previous moment. Considering the displacement can be in any direction, this head is activated by a hyperbolic tangent function, making the range of values output -1 and 1.

Figure 6 presents the model's complete output, including object centers, dimensions, and displacements. These displacements are essential for associating objects to specific trajectories, enabling effective tracking over time.

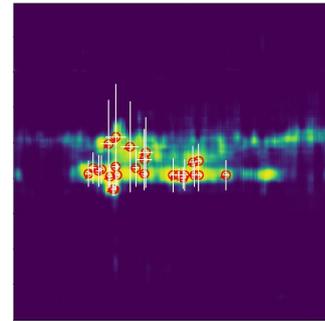

**Figure 6.** Complete model output image. Here you can see the size of each detected object (white bars) and the movement direction (red arrows) from their centers.

### 3.5 Post-Processing and Data Association

Each model head represents a piece of the complete MOT output. However, two additional steps are required to fully build the tracking routine. The first involves processing and uniting all model outputs, while the second employs a common approach to associate current inferences with the object's tracking history.

#### 3.5.1 Combining Outputs

To integrate the model outputs, a specific procedure is followed:

1. **Apply a Threshold on the Heatmap:** This step identifies the detected object centers by applying a threshold to the heatmap.
2. **Determine Object Dimensions and Displacements:** Using the identified centers, the object's dimensions and displacements are determined.

In Figure 5 we see the raw outputs of each detection found by the model in a given frame. It is to possible to see that some of these detections are either overlapping or are not precise. This is addressed by applying the Non-Maximum Suppression (NMS) algorithm to mitigate false positives, by suppressing overlapping and low-confidence detections. This method is widely used in object detection systems, including YOLO variants such as Wang *et al.* [2022].

Given a set of bounding boxes $B$ with their respective confidence scores $S$ and an overlap threshold $T$, the NMS algorithm proceeds as follows:

1. **Select the Box with the Highest Confidence Score:** Select bounding box $b_{max}$ with the highest confidence score $s_{max}$ from $S$.

2. **Compare with Other Boxes:** Compare $b_{max}$ with each box $b_i$ in BB, except itself.
3. **Calculate Intersection over Union (IoU):** For each comparison, calculate the Intersection over Union (IoU) between $b_{max}$ and $b_i$, denoted by $IoU(b_{max}, b_i)$.
4. **Suppression:** If $IoU(b_{max}, b_i) > T$, box $b_i$ is considered redundant and removed from $B$.
5. **Repeat:** Repeat steps 1 to 4 until all boxes in $B$ are evaluated or removed.

Figure 7 illustrates predictions before and after NMS, demonstrating the algorithm's effectiveness in reducing false positives. Image (a) shows detected objects immediately after the image is processed by the model, while image (b) shows after the redundancy reduction algorithm.

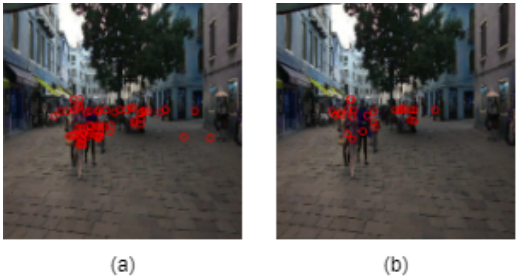

**Figure 7.** Image (a) shows the detections over the image before applying NMS step. Image (b) shows how the detections become after applying NMS.

To track detected objects, current detections must be associated with previous ones. The Hungarian algorithm is an optimization method for solving assignment problems. Its goal is to find the perfect match with the minimum cost between elements of two sets, minimizing the total association cost. In the context of multi-object tracking, it is used to associate current detections with existing trajectories based on a distance or similarity metric (in this case, the IoU between bounding boxes).

Formally, the Hungarian algorithm solves the assignment problem as follows: Let $C$ be a cost matrix $n \times m$, where $n$ is the number of detected objects and $m$ the number of trajectories. The algorithm seeks a permutation $P$ of $m$ elements such that the sum of assignment costs $\sum_{i=1}^{n} C_{i,P(i)}$ is minimized, where $P(i)$ is the trajectory assigned to detected object $i$.

Given a set of detected objects $D$ and a set of existing trajectories $T$, the algorithm proceeds as follows:

1. **Construct the Cost Matrix:** A cost matrix $C$ is built, where each element $C_{i,j}$ represents the cost of assigning detected object $d_i$ to trajectory $t_j$. The cost can be defined based on Euclidean distance, inverse IoU, or any other relevant dissimilarity metric.
2. **Apply the Hungarian Algorithm:** The algorithm is applied to the cost matrix to find the minimum-cost assignment. This results in a one-to-one mapping between detected objects and trajectories, where each object is associated with at most one trajectory and vice versa.
3. **Update Trajectories:** Based on the assignments made, trajectories are updated with new detections. Trajectories without corresponding detections may be marked as "lost" or terminated. Similarly, detections without corresponding trajectories are marked as new objects.

### 3.6 Training Configurations

#### 3.6.1 Part-Based Multitask Training

As discussed in section 2, Multi-Object Tracking (MOT) models are often trained end-to-end using a multitask learning strategy. However, the results presented in this study, consistent with observations from FairMOT Zhang *et al.* [2021], indicate that simply assigning a specific loss function to each task does not guarantee optimal model performance. In response to this limitation, a training methodology was adopted that alternates between phases of individualized task learning and joint training phases, known as Part-Based Multitask Training (TMP).

The training routine follows this process, called Part-Based Multitask Training (TMP):

1. **Heatmap Head Training Phase:** Initially, the heatmap generation task is isolated, freezing the weights of the dimension and displacement heads and nullifying their respective weights in the weighted combination of loss functions. This stage focuses exclusively on learning object centers.
2. **Training Period:** This phase lasts for 50 epochs or continues until no reduction in the loss function is observed over 10 epochs.
3. **Individual Training of Other Heads:** The process is repeated for the dimension and displacement heads, training each individually with the same criteria for weight freezing and duration.
4. **Final Joint Training:** After completing 150 focused training epochs, a final joint training phase of 50 epochs is conducted, where all layers are activated and contribute their weights to the weighted average of the loss functions.

This sequential approach allowed the model to learn the specifics of each task in isolation before integrating that knowledge into joint learning. This process facilitated the model's inter-task understanding, resulting in superior performance at the end of the training.

#### 3.6.2 Loss Functions

During the training of the Multi-Object Tracking (MOT) model, various specific loss functions were applied to improve the model's accuracy in detecting and tracking objects. These loss functions are essential to guide the model in accurately detecting object centers as well as their dimensions and displacements. For clarity, each implemented loss function is detailed below and expressed through mathematical equations.

**Heatmap Loss** The heatmap loss function is designed to optimize the detection of object centers in the heatmap. This function consists of two main parts: **Center Loss** and **Focal Loss**.

**Center Loss** is calculated as follows:

$$C_{loss} = -\frac{1}{N}\sum_{i=1}^{N}\left[P_i \cdot \log(\hat{P}_i) + (1-P_i) \cdot \log(1-\hat{P}_i)\right] \cdot W_i \quad (1)$$

where $P_i$ is the ground truth probability of a pixel being an object center, $\hat{P}_i$ is the predicted probability, $N$ is the total number of pixels, and $W_i$ represents weights determined by the importance of each pixel. This loss function is the same as that demonstrated by CenterTrack Zhou *et al.* [2020].

**Focal loss** is defined as:

$$\text{FocalLoss} = -\frac{1}{N}\sum_{i=1}^{N}(1-\hat{P}_i)^{\gamma} \cdot P_i \cdot \log(\hat{P}_i) \quad (2)$$

where $\gamma$ is the parameter that modulates the loss for well-classified examples, focusing learning on misclassified examples. In this model, $\gamma$ is set to 4.

**Grid Loss** This function optimizes the grids representing object dimensions and displacements. It is based on the L1 loss between predictions and ground truth values, applied only where objects are present (mask):

$$\text{GridLoss} = \frac{1}{M}\sum_{i=1}^{M}|G_i - \hat{G}_i| \quad (3)$$

where $G_i$ are the actual grid values, $\hat{G}_i$ are the values predicted by the network, and $M$ is the number of elements selected by the mask.

These loss functions allow the model to accurately learn object location, dimensions, and displacements, facilitating effective tracking in image sequences. They are combined using a weighted average. In the equation below, we see how the final loss value is calculated for each image batch:

$$Loss_f = \frac{(CenterLoss + FocalLoss)*w1}{w1+w2+w3} + \\ \frac{(GridLoss_{dim})*w2}{w1+w2+w3} + \quad (4) \\ \frac{(GridLoss_{dlc})*w3}{w1+w2+w3}$$

where $w1 = w2 = w3 = 1.0$ for the final training. As described in TMP, these weights vary to 0 depending on the training stage. $GridLoss_{dim}$ refers to GridLoss applied to the dimension output, and $GridLoss_{dlc}$ to GridLoss applied to the displacement output.

## 4 Results

This section details the experiments conducted and the results obtained, providing insights into the performance of the OneTrack-M model.

### 4.1 Experimental Setup

The experiments were conducted on a computer equipped with an Intel Core i7-11700 processor at 3.60GHz and 32GB of RAM, running the Ubuntu 20.04.6 LTS 64-bit operating system. For the training and inference of the model, a Geforce RTX 2060 graphics card with CUDA 11.0 support was used. The choice of the graphics card was based on the compute capability score provided by the manufacturer, ensuring a computing capacity comparable to the Nvidia V100, commonly used in related work. This approximation ensures the validity of the comparison of inference times with previously published results.

### 4.2 Dataset

The MOT17 benchmark Milan *et al.* [2016] was the dataset used to validate the results of this work. It was chosen because it is widely adopted in MOT research to maintain a consistent comparative baseline. MOT17 includes predefined training and test datasets. The images cover various scenarios, varying in object proximity and camera movement, focusing on tracking people in moderately dense environments. Figure 8 exemplifies a frame from the dataset, which contains 18 annotated image sequences with bounding box coordinates and object identifications.

MOT17 contains a total of 18 image sequences with annotations of bounding box coordinates and IDs within the sequence. The annotations follow a format of top, left, height, and width for each object, as well as its prediction confidence and the respective ID (new or not). Figure 8 shows an example of a sample from this dataset, containing a sequence of 3 examples separated by 60 frames.

### 4.3 Evaluation Metrics

A set of metrics based on two standards established in the literature was adopted: Clear MOT and HOTA. Clear MOT, detailed in Bernardin and Stiefelhagen [2008], approximates the tracking challenge to concepts such as accuracy, precision, and recall, while HOTA, introduced in Luiten *et al.* [2020], focuses on specific aspects of MOT, such as consistent trajectory maintenance. The TrackEval library, by Jonathon Luiten [2020], was used to evaluate the model outputs according to these metrics.

The metrics used are:

- **MOTA (Multiple Object Tracking Accuracy)**: An aggregate measure that considers false positives, false negatives, and identification errors to calculate overall tracking accuracy. It is expressed as:

$$\text{MOTA} = 1 - \frac{\sum_t(FN_t + FP_t + IDSW_t)}{\sum_t GT_t} \quad (5)$$

where $FN_t$, $FP_t$, and $IDSW_t$ represent the number of false negatives, false positives, and identification switches at each time step $t$, respectively, and $GT_t$ is the total number of objects.

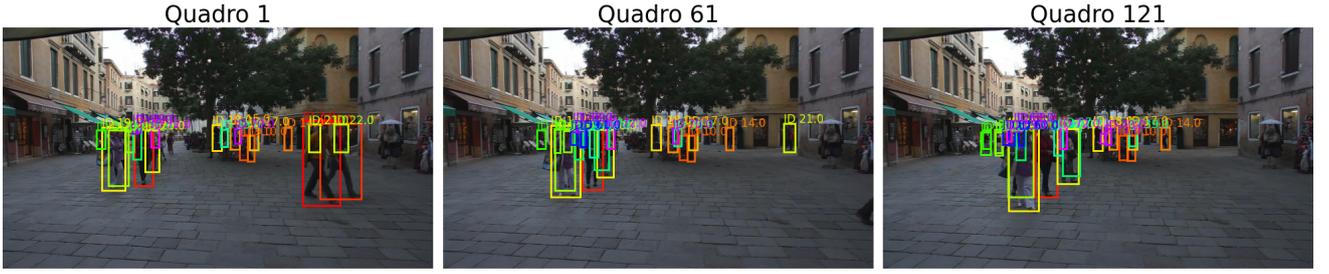

**Figure 8.** Image sequence and example annotations taken from the MOT17 dataset.

- **MOTP (Multiple Object Tracking Precision)**: Measures tracking precision by calculating the average distance between predicted and actual object locations. It is defined as:

$$\text{MOTP} = \frac{\sum_{i,t} d_{i,t}}{\sum_t C_t} \quad (6)$$

where $d_{i,t}$ is the distance between the predicted and actual location of the object $i$ at time $t$, and $C_t$ is the total number of detected matches.

- **HOTA (Higher Order Tracking Accuracy)**: Combines detection and association capabilities into a single metric, balancing performance in both dimensions.

$$\text{HOTA} = \frac{1}{T} \sum_{t=1}^{T} \frac{\sum_{(i,j) \in \mathcal{M}_t} \text{IOU}(i,j)}{|\mathcal{M}_t|} \cdot \frac{|\mathcal{M}_t|}{|\mathcal{G}_t| + |\mathcal{P}_t| - |\mathcal{M}_t|}) \quad (7)$$

In this equation, $\mathcal{M}_t$ represents the set of matched ground truth-prediction pairs at time $t$, $\mathcal{G}_t$ is the set of ground truth objects, $\mathcal{P}_t$ is the set of predicted objects, and $\text{IOU}(i,j)$ is the Intersection over Union between the ground truth object $i$ and the predicted object $j$. $T$ is the total number of time steps.

- **IDS (ID Switches)**: Quantifies the number of times an object's identification changes over its trajectory. It is an indicator of the consistency in maintaining object identities.

- **FPS (Frames per second)**: This metric relates to a term used to quantify video capture speed. In the context of this work and measurements of inference times for video processing models, this metric defines how many frames per second the model can infer, i.e., a frequency measure, being the inverse of the model's average inference time for a given sequence:

$$\text{FPS} = \frac{1}{I_t/N} \quad (8)$$

where $I_t$ is the total inference time of a sequence in seconds and $N$ is the number of frames in this sequence.

These metrics provide a comprehensive evaluation of the model's performance, from object detection accuracy to the effectiveness in maintaining their trajectories and identities over time. In addition to these metrics, the model's average inference time on this dataset was also evaluated, measured from the number of frames processed per second, or FPS. This measure includes the preprocessing, inference, post-processing, and association process.

### 4.4 Benchmark Test Results

Table 2 shows the results obtained by the model compared to other established models in the literature.

The main benefit of this approach is its inference speed. However, it was able to outperform the CenterTrack model, which becomes a baseline regarding the use of centers and displacements to perform MOT. Conversely, the MOTR model is another important work as it demonstrates the application of transformers to MOT. The OneTrack-M model proposed in this work improved on both concepts, surpassing these baselines in all metrics, especially in FPS.

Now comparing with other works using different approaches, we see that there are still places to be improved. In terms of MOTA, our model shows a rather big discrepancy from most of the other works, with up to 12.65% difference. This is compensated with the other metrics, but it is still interesting to understand why this single metric is lacking. For HOTA, we achieve a 65.105%, which is around 0.5% higher than the second highest performing model, BoTSORT. A high HOTA suggests that both detection and association processes are performing well. For IDF1, our model configuration achieves a 79.608%, 0.108% higher than the second highest performing models, tied at 79.5% for both StrongSORT and BoTSORT. A high IDF1 indicates a strong performance in matching detected objects to their true identities. A low number of ID switches (IDS) implies that the tracking algorithm is good at maintaining the identity of tracked objects, meaning the tracker is reliable in associating the correct IDs across frames. For this test, we have around 100 fewer occurrences of ID switches in comparison to the second highest, StrongSORT, at 1194. Finally, FPS represents how fast the model runs its inferences. This is the strongest suit for our work, as the models that achieve higher metrics, tend to take a big hit to how fast it can run. At 35.7 FPS, our model would be able to run in real-time considering footage capturing 30 FPS in a practical scenario.

False positives and false negatives in the detection step can have a high impact on MOTA, while the others do not so much care for false tracking. This indicates that the model can effectively track objects it finds, but it is seeing some either wrong or redundant detections, affecting the MOTA metric, but this is not taken into account by the other metrics, they give more value to correct trackings.

The configuration selected for this result was considered the most balanced between performance on metrics and the lowest inference time. In this case, 5 images per inference, with TMP and using the ViT-Base-16 as the backbone.

Table 2. Comparison of Results between established models in the literature on the MOT17 dataset

| Model | HOTA | MOTA | IDF1 | IDS | FPS |
|---|---|---|---|---|---|
| **OneTrack-M** | **65.105** | 67.950 | **79.608** | 1090 | **35.7** |
| Bytetrack | 63.1 | 80.3 | 77.3 | 2196 | 11.8 |
| CenterTrack | 52.2 | 67.8 | 64.7 | 3039 | 17.5 |
| MOTR | - | 65.1 | 66.4 | 2049 | 7.5 |
| MOTRV2 | 62.0 | 78.6 | 75.0 | - | 6.9 |
| TrackFormer | - | 74.1 | 68.0 | 2829 | 7.4 |
| TransTrack | 54.1 | 75.2 | 63.5 | 3603 | 10.0 |
| BoTSORT | 64.6 | 80.6 | 79.5 | 1257 | 6.6 |
| StrongSORT | 64.4 | 79.6 | 79.5 | 1194 | 7.1 |
| FairMOT | - | 73.7 | 72.3 | 3303 | 25.9 |
| LMOT | - | 72.0 | 70.3 | 3071 | 28.5 |

These results mainly highlight the model's ability to maintain trajectories, i.e., recognize which objects are correlated over time, demonstrating that the mechanisms developed for this during the model's conception were relevant to these results. Another relevant point is the processing time, which, despite using a transformer model, by performing other steps efficiently and simplifying the process in a single pass, reported the best inference time among the models considered for this work. This version of the model presents the following configurations: using the TMP method, Vision Transformer version Base-16, and a window of 5 images at a time. This proved to be the most stable configuration, as it balances processing time, qualitative performance, and the ability to correlate relevant information in the time window for each object.

## 4.5 Model Characteristics Test Results

These tests aim to validate some of the decisions made for the final version of the model. In this part of the work, we evaluate different window sizes, training with and without TMP, different configurations for the feature extractor, and the application of a usual positional embedding (like the one used in ViT Dosovitskiy *et al.* [2021]) or the proposed Channel Wise Encodings to provide context for each crop in the sequence of the transformer network.

### 4.5.1 Window Size Test

In Table 3, it can be seen that processing time is not affected by the adopted window size. This makes sense and it is in line with how the model processes its inputs, stacking all window images along the RGB channel dimension, effectively using more memory to process more images per window. However, it is not advisable to increase this parameter uncontrollably, as more frames make it more difficult for the model to predict displacement, mainly because this displacement should be from the previous frame's position. An increase in ID values is noticeable when receiving context from time instances far in the past. Thus, a value that keeps this error rate acceptable was 5.

Table 3. Table of results on window size.

| Window Size | HOTA | MOTA | IDF1 | IDS | FPS |
|---|---|---|---|---|---|
| 1 | 45.291 | 45.822 | 56.379 | 5145 | 35.7 |
| 2 | 56.159 | 58.944 | 74.388 | 1449 | 35.7 |
| 3 | 56.316 | 61.813 | 76.969 | 1365 | 35.7 |
| 4 | 62.05 | 64.305 | 78.41 | 1253 | 35.7 |
| 5 | 65.105 | 67.950 | 79.608 | 1090 | 35.7 |
| 10 | 52.525 | 56.821 | 64.061 | 3095 | 35.7 |
| 20 | 52.409 | 56.627 | 63.919 | 3193 | 35.7 |

### 4.5.2 TMP Effectiveness Test

To evaluate the capability of this training method, two versions of the model were trained, one training all loss functions at once, as normally done, for 200 epochs. The other version was trained following the training routine proposed by TMP, training each head or task of the model individually, freezing the weights of the other heads, one at a time, followed by a final training with all heads at once. Table 4 shows the results of each test. We can see that the model greatly benefits from an isolated training routine for each task. For this test, all other model hyperparameters were kept constant, i.e., window size fixed at 5 images at a time, and the encoding step by ViT uses the B16 model, i.e., the base size of the network, with images cropped to 16x16 pixels.

Table 4. Comparison of results between a model trained with and without the TMP technique.

| With TMP? | HOTA | MOTA | IDF1 | IDS | FPS |
|---|---|---|---|---|---|
| No | 49.915 | 55.398 | 72.371 | 2312 | 37.7 |
| Yes | 65.105 | 67.950 | 79.608 | 1090 | 37.7 |

### 4.5.3 Test of Different Transformer Feature Extractors

One of the essential parts for the good performance of this model is the choice of the architecture responsible for feature extraction. This is because architecture requires this step to understand concepts in both the spatial and temporal axes. Therefore, this part must be robust enough to handle these aspects simultaneously.

This also brings a relevant question to the overall inference time of the system. This model part is responsible for almost all of the pipeline's time consumption. Thus, applying a network with many parameters greatly impacts the model's processing time per second. This is well reflected in the obtained results, shown in Table 5.

For more details on the configuration of each ViT version, they should be referenced from the original work in Dosovitskiy *et al.* [2021], but briefly, the main difference for the Large models is a larger number of parameters and layers, while the 16 and 32 number in models refer to the size at which the images are cropped. This implies that the 32 models have less detail representation capability as they need to summarize more information in each input token to the encoder sequence. Conversely, this larger size per crop means fewer elements per image, resulting in a shorter inference time.

**Table 5.** Comparison results of the model varying the version of the Vision Transformer being used.

| ViT Version  | HOTA   | MOTA   | IDF1   | IDS  | FPS   |
|--------------|--------|--------|--------|------|-------|
| ViT-Base-16  | 65.105 | 67.950 | 79.608 | 1090 | 37.7  |
| ViT-Base-32  | 60.310 | 65.466 | 71.230 | 1322 | 60.18 |
| ViT-Large-16 | 68.019 | 75.020 | 80.052 | 1108 | 8.11  |
| ViT-Large-32 | 65.019 | 71.020 | 81.166 | 1243 | 21.88 |

#### 4.5.4 Contextual Embeddings Test

This test aims to validate the method presented in this work regarding the embedding that incorporates spatial and temporal context. During preprocessing, the image sequence is stacked in the channel dimension, and applying these representations to each channel promotes spatial-temporal perception by the model. A common practice for this part of transformer models includes using spatial embeddings, either through fixed sinusoidal functions that set values for each position in the original image or through learnable representations with weights adjusted during training. The results of this test are presented in Table 6.

**Table 6.** Comparison results of the model varying the embedding method used.

| Embedding Method | HOTA   | MOTA   | IDF1   | IDS  | FPS  |
|------------------|--------|--------|--------|------|------|
| Positional       | 44.678 | 45.207 | 58.223 | 4866 | 39.5 |
| Channel Wise     | 65.105 | 67.950 | 79.608 | 1090 | 37.7 |

The results indicate that the model does not identify temporal features when handling input data, and only positional embedding is used. Even with an inference window of 5 images, there are difficulties in properly connecting the movement of objects in the scene, resulting in tracking failures, as shown by the high incidence of identification switches.

### 4.6 Qualitative Analysis Results

Figure 9 shows a sequence of images with their respective inferences. It demonstrates a problem called ID theft. Due to the method used to associate IDs over time, this phenomenon becomes present in the inferences, where a sufficiently large error in the displacement value for two very close objects can cause one of them to receive the ID of the other, while it becomes necessary to create a new ID for the one that lost its identifier.

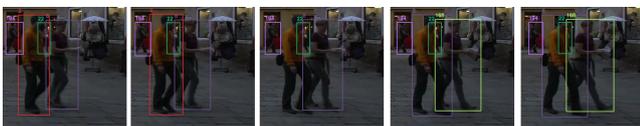

**Figure 9.** Example of an ID theft case. Note that between the second and fourth image, the object with id=2 loses it and receives the ID of the object next to it as a new value.

Next, Figure 10 demonstrates a model limitation for objects at medium to long distances from the camera. Here we can see a person on a bicycle that is not detected over several frames while other elements appear to be positioned close to the person in a two-dimensional notion, however, due to the perspective generated by the image, the model does not detect them.

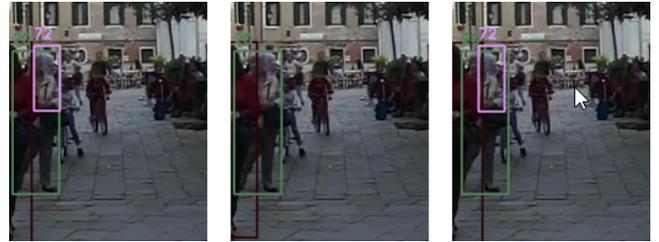

**Figure 10.** Example of an object too far for detection and tracking.

Another interesting case, demonstrating the model's ability to maintain the trajectory despite occlusions, is in Figure 11. In this image, there is a person completely behind the person in front of the image. In the second frame, the model even loses the detection of this person, but in the last one, the ID is recovered, as the same one that had been lost, despite the case showing severe occlusion.

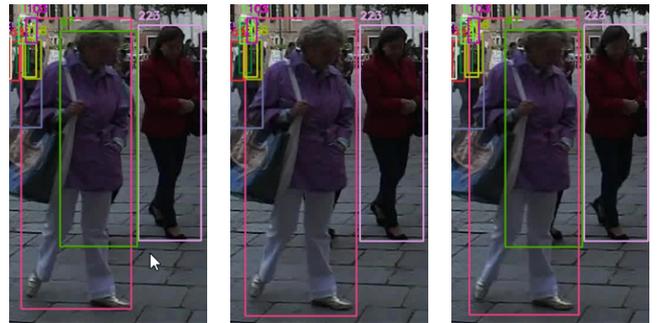

**Figure 11.** Demonstration of model robustness in a scenario of object occlusion.

In summary, while the model can show some weak points, which can and should be considered for future works, it is compensated in higher metrics (besides MOTA) and especially in high FPS, which can allow for better practical usage of this model, which was not available with some of the better performing models, scoring less than 10 FPS, meaning in a 30 FPS video, it would be missing a third of frames for processing.

## 5 Conclusion and Future Works

This work brings relevant contributions to the field of object tracking. These innovations focused on improving two main aspects: inference time and training stability. Regarding inference time, the use of transformers allowed the abstraction of the temporal context in such a way that the network is responsible for relating the parts of the network, without requiring an internal step of temporal contextualization. That is, instead of modeling the problem as each element of the transformer's input sequence being an image from the frame window, they all become a single image with multiple channels. From the implemented encoding, which provides these channels with temporal information, the model can consider everything as a single image. In addition, there is no need for a decoder to infer objects and their trajectories, requiring only a few more layers at the output to process the generated attention maps.

In terms of training stability, the TMP method proved to be an option with the potential to improve the training of MOT models in general. Making the process more stable and generally improving the results obtained.

Finally, as future work, we suggest investigating the feasibility of TMP for other MOT models that have similar processing, that is, perform both MOT stages - detection and tracking - in a unified way. This is also true in other fields where there are models that also perform related tasks but may end up interfering with the mutual learning of the model.

Additionally, specifically for OneTrack-M, evaluating other extractor models instead of the Vision Transformer may be a promising path, but it requires additional work to change the functioning of the proposed architecture mechanisms in such a way that it maintains its characteristics and allows effective training of the model.

# Declarations

## Funding

No monetary funding was provided for the development of this work.

## Authors' Contributions

All authors contributed equally to this work. All authors read and approved the final manuscript.

## Competing interests

The authors declare that they have no competing interests.

## Availability of data and materials

The datasets generated and/or analyzed during the current study are available on the MOTChallenge website, specifically at https://motchallenge.net/data/MOT17/.